%% file: main.tex
\documentclass[10pt,twocolumn,letterpaper]{article}

\usepackage[pagenumbers]{cvpr}

\input{preamble}

\definecolor{cvprblue}{rgb}{0.21,0.49,0.74}
\usepackage[pagebackref,breaklinks,colorlinks,allcolors=cvprblue]{hyperref}

\def\confName{CVPR}
\def\confYear{2026}

\title{Grounding Video Reasoning in Physical Signals}

\author{
Alibay Osmanli \qquad Zixu Cheng \qquad Shaogang Gong\\
Queen Mary University of London\\
{\tt\small \{a.osmanli,zixu.cheng,s.gong\}@qmul.ac.uk}
}

\begin{document}
\maketitle

\input{sec/0_abstract}
\input{sec/1_intro}
\input{sec/2_related}
\input{sec/3_method}
\input{sec/4_dataset}
\input{sec/5_experiments}
\input{sec/6_results}
\input{sec/7_limitations}
\input{sec/8_conclusion}

{
    \small
    \bibliographystyle{ieeenat_fullname}
    \bibliography{main}
}

\end{document}

%% file: preamble.tex
\usepackage{amsmath,amssymb}
\usepackage{booktabs}         
\usepackage{multirow}         
\usepackage{graphicx}
\usepackage{xcolor}
\usepackage{microtype}
\usepackage{enumitem}
\usepackage{comment}

%% file: sec/0_abstract.tex
\begin{abstract}
Physical video understanding requires more than naming an event correctly. A model can answer a question about pouring, sliding, or collision from textual regularities while still failing to localize the event in time or space. We introduce a grounded benchmark for physical video understanding that extends the \textit{what--when--where} evaluation structure of V-STaR~\cite{vstar} to four video sources, six physics domains, three prompt families (\texttt{physics}, \texttt{vstar\_like}, and \texttt{neutral\_rstr}), and four input conditions (original, shuffled, ablated, and frame-masked). The benchmark contains 1,560 base video clips from SSV2~\cite{ssv2}, YouCook2~\cite{youcook2}, HoloAssist~\cite{wang2023holoassist}, and Roundabout-TAU~\cite{lin2026taur1}. Each clip is first converted into a shared grounded event record, and the three query families are derived from that record. Temporal and spatial targets are shared across prompt families, while the non-physics families use deterministic family-appropriate semantic \texttt{a\_what} targets derived from the same record. Across models and prompt families, \texttt{physics} remains the strongest regime overall, \texttt{vstar\_like} is the clearest non-physics semantic comparison, and \texttt{neutral\_rstr} behaves as a harder templated control. Prompt-family robustness is selective rather than universal, perturbation gains cluster in weak original cases, and spatial grounding is the weakest across settings. These results suggest that video Q$\&$A reasoning benchmarks shall report physically grounded, prompt-aware, and perturbation-aware diagnostics alongside aggregate accuracy.
\end{abstract}

%% file: sec/1_intro.tex
\section{Introduction}

In a video question answering task, a correct answer to a physical video question does not guarantee visually grounded understanding. A model can answer a question about pouring, sliding, or collision because the wording narrows the event type, even if it never identifies when the event happens or where the relevant objects are in a video. This matters in physical scenes, where motion, contact, force, and state change unfold over time and are often spatially localized. Standard answer-only benchmarks do not separate these cases. They tell us whether the final response matches the label, not whether the model uses video in a grounded~way~\cite{buch2022revisiting}.

Recognising that a clip contains a ``collision'' from the question text is different from locating the interaction in time and grounding the participating objects in space. In robotics, video moment retrieval, and embodied decision-making, the first kind of success is not enough.

Existing benchmarks cover only part of this problem. Physical reasoning datasets such as CLEVRER~\cite{clevrer}, IntPhys~\cite{intphys}, and PhysBench~\cite{chow2025physbench} probe physical understanding, but they mostly rely on categorical outputs. Grounded video benchmarks~\cite{zhang2020vidstg,tvqaplus,nextgqa} evaluate temporal or spatial localisation, but they are not organized around physical event structure. V-STaR~\cite{vstar} took an important step by showing that strong performance on \emph{what} can coexist with weak performance on \emph{when} and \emph{where}. That raises a useful diagnostic question for physical video understanding: when a model answers correctly, is it actually grounded in the video?

We study this question by introducing a physics-focused benchmark. Our benchmark keeps the grounded \textit{what--when--where} structure of V-STaR, but moves it to six physics domains: Gravity, Fluids, Collisions, Deformation, Friction, and State Changes. Each sample is organized around a shared grounded event record with an event description, temporal span, and bounding box. The three prompt families---\texttt{physics}, \texttt{vstar\_like}, and \texttt{neutral\_rstr}---are alternative query formulations derived from that same record, and each is evaluated under four input conditions---original, shuffled, ablated, and frame-masked. The \texttt{physics} family is the main benchmark regime, \texttt{vstar\_like} is a V-STaR-style semantic comparison rather than an exact reconstruction of the original benchmark prompts, and \texttt{neutral\_rstr} is a neutral wording control that functions as a templated ablation.

The grounded event record stays fixed across prompt families, but the semantic field is expressed differently across them. Both \texttt{a\_when} and \texttt{a\_where} are scored against shared targets throughout. For \texttt{neutral\_rstr} and \texttt{vstar\_like}, \texttt{a\_what} is scored against shorter family-appropriate semantic targets derived from the same record. Under this protocol, \texttt{physics} remains the strongest overall regime, \texttt{vstar\_like} is the main non-physics comparison, and \texttt{neutral\_rstr} remains the harder control. These shifts are not uniform across models. Some remain comparatively strong across formulations, while others depend more heavily on the cues made explicit by physics framing.

\begin{figure*}[ht!]
\centering
\includegraphics[width=1\textwidth]{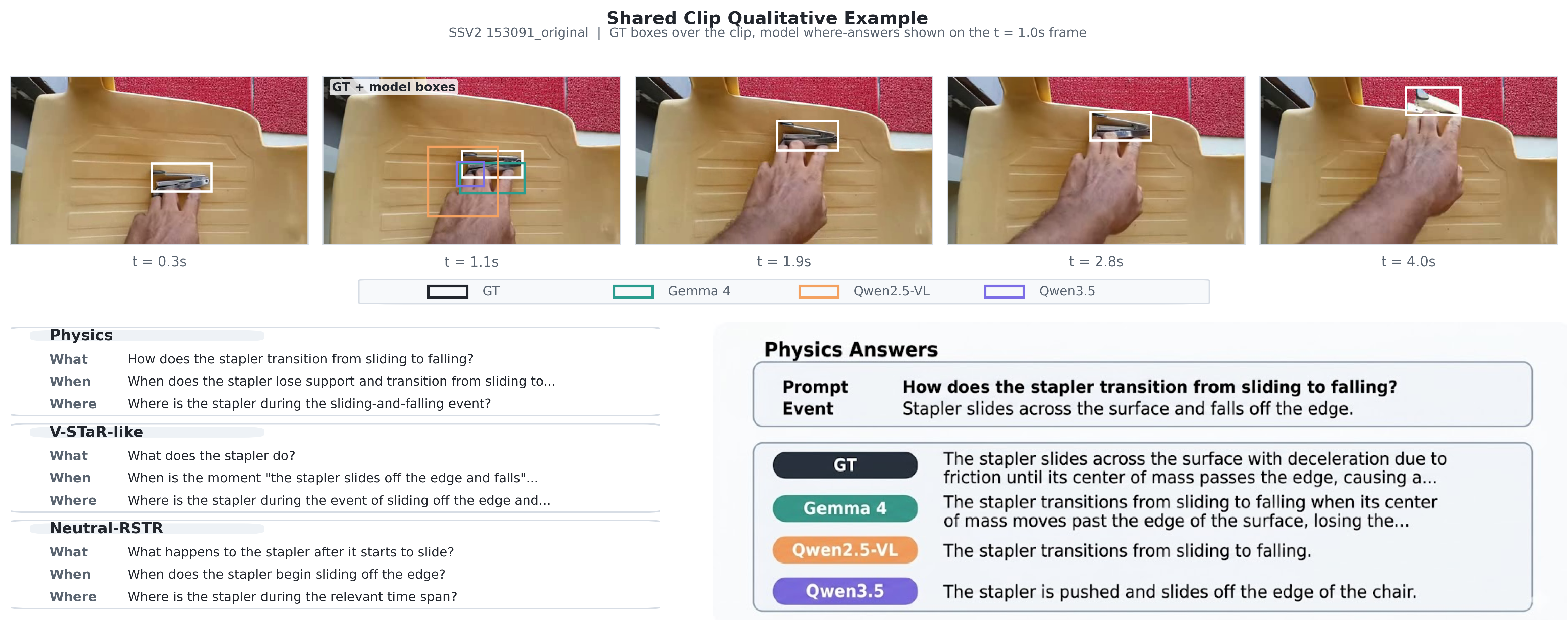}
\caption{One shared clip across prompt families and grounded outputs. Top: matched SSV2 frames with the ground-truth box trajectory and representative model \texttt{a\_where} predictions. Bottom left: the three query families derived from the same event record. Bottom right: representative physics answers from the ground truth and three models.}
\label{fig:prompt_families}
\end{figure*}

Aggregate scores still hide much of this behavior. Two models with similar overall performance can react very differently when the wording changes or the visual evidence is perturbed.
Figure~\ref{fig:prompt_families} previews the setup on one shared clip. The prompt families ask about the same event, but the ground-truth and model box overlays show that plausible answers can still hide very different spatial grounding.

\paragraph{Research questions.}
We organize this study around four questions. RQ1 asks whether the grounded failure pattern identified by V-STaR persists in physical video understanding. RQ2 asks how stable grounded performance remains across \texttt{physics}, \texttt{vstar\_like}, and \texttt{neutral\_rstr} when the underlying event record is fixed. RQ3 asks what perturbation gains and losses reveal beyond original-condition performance. RQ4 asks which weaknesses remain even for strong models and favorable prompt families.

We keep the hypotheses narrow. We expect the V-STaR failure pattern to persist in physics-focused video, with spatial grounding remaining the weakest component. We also expect \texttt{physics} to be strongest overall, \texttt{vstar\_like} to be the strongest non-physics comparison, and \texttt{neutral\_rstr} to behave as a harder templated control. Finally, we expect perturbation gains to concentrate in weak or mid-baseline cases and to function as diagnostics of evidence sensitivity rather than as simple robustness wins.

\noindent Our main contributions are:
\begin{itemize}
\item A grounded benchmark for physical video understanding, built from 1,560 base clips from SSV2~\cite{ssv2}, YouCook2~\cite{youcook2}, HoloAssist~\cite{wang2023holoassist}, and Roundabout-TAU~\cite{lin2026taur1}, and organized into six physics domains.
\item A three-prompt evaluation design built on the same shared event record, separating the main \texttt{physics} regime from a V-STaR-style prompt family and a neutral wording control.
\item A perturbation analysis framework that combines shuffled, ablated, and frame-masked inputs with component-wise metrics and diagnostic indices to interpret changes under degraded evidence.
\item Empirical evidence that prompt-family robustness is selective across models, perturbation gains concentrate in weaker original cases, and spatial grounding remains the most persistent weakness across model families.
\end{itemize}

%% file: sec/2_related.tex
\section{Related Work}

\subsection{Video-LLM evaluation}

Recent video-LLMs combine a visual encoder with a pretrained language model and instruction tuning over mixed-modality datasets~\cite{maaz2024videochatgpt,lin2023videollava}. Systems such as Qwen2.5-VL~\cite{qwen25vl}, Qwen3-VL~\cite{bai2025qwen3vl}, VideoLLaMA3~\cite{videollama3}, InternVideo2.5~\cite{internvideo25}, InternVL3.5~\cite{wang2025internvl35}, and MiniCPM-o~\cite{minicpm} perform well on standard benchmarks including Video-MME~\cite{videomme}, MVBench~\cite{mvbench}, and LongVideoBench~\cite{longvideobench}. Those benchmarks are useful for broad coverage, but they score the final answer rather than the grounding process. A correct answer does not tell us whether the model located the relevant event in time and space or whether it relied on prompt cues and dataset regularities.

\subsection{Physical reasoning benchmarks}

Physical reasoning has been studied in synthetic and real-world settings for a long time. IntPhys~\cite{intphys} evaluates physical plausibility, while CLEVRER~\cite{clevrer} probes causal and counterfactual reasoning in a controlled collision world. More recent benchmarks such as PhysBench~\cite{chow2025physbench} evaluate multimodal LLMs on physical concepts including gravity, collision, and material behavior. The common limitation is the answer format. Most of these benchmarks use categorical or multiple-choice outputs, so a model can score well without grounding the event itself in time or space.

\subsection{Grounded video understanding}

Grounded video understanding is usually studied through temporal grounding, spatio-temporal grounding, and grounded question answering. TALL and Charades-STA formalized temporal moment localization in untrimmed video~\cite{gao2017tall}, while ActivityNet Captions extended that setting to dense event description~\cite{krishna2017dense}. VidSTG~\cite{zhang2020vidstg} made spatio-temporal grounding a standard evaluation problem, and TVQA+~\cite{tvqaplus} and NExT-GQA~\cite{nextgqa} added temporal and spatial annotations to video Q$\&$A. These datasets make grounding visible, but they are not designed around physical event structure or prompt-sensitive diagnosis.

\subsection{Diagnostic evaluation and V-STaR}

Several recent works argue that strong aggregate scores can conceal brittle behavior. Buch~\etal~\cite{buch2022revisiting} showed that some video-language benchmarks can be solved from linguistic structure alone. Bagad~\etal~\cite{bagad2023testoftime} found that leading video-language models remain weak at chronological reasoning when frame order is manipulated. In other modalities, Winoground~\cite{thrush2022winoground} and CheckList~\cite{ribeiro2020beyond} showed the value of controlled diagnostic testing instead of relying only on headline benchmark scores.

V-STaR~\cite{vstar} is the closest prior benchmark to this study. It introduced a grounded \textit{what--when--where} evaluation structure and showed that video-LLMs often perform much better on \emph{what} than on \emph{when} or \emph{where}. Our benchmark uses the same diagnostic logic. Against physical reasoning benchmarks, we add grounded outputs. Against standard grounding benchmarks, we add a physics-domain layer and controlled perturbations. We also include a V-STaR-style prompt family explicitly, not leaving that comparison implicit.

%% file: sec/3_method.tex
\section{Benchmark design}

\subsection{Task format and prompt families}

Our benchmark inherits the grounded \textit{what--when--where} evaluation structure introduced by V-STaR~\cite{vstar}. Given a video clip and a physical question, the model must produce a single structured prediction with three fields:
\begin{itemize}
\item \texttt{a\_what}: a short text description of the physical event,
\item \texttt{a\_when}: a temporal interval [\texttt{start\_sec}, \texttt{end\_sec}],
\item \texttt{a\_where}: a normalised bounding box \texttt{x}, \texttt{y}, \texttt{w}, and \texttt{h}.
\end{itemize}

Requiring all three parts in one response is deliberate. Recognition, temporal grounding, and spatial grounding can each be gamed differently if they are evaluated in isolation. By forcing the model to commit to one coherent account of the event, we make it harder to score well by naming the event while ignoring where it happened or when it started.

We evaluate the same grounded event record under three prompt families:
\begin{itemize}
\item \textbf{\texttt{physics}}: the main benchmark regime, written to foreground physical dynamics and event descriptions.
\item \textbf{\texttt{vstar\_like}}: questions written in the style of the original V-STaR prompts using the existing annotations only. It is a continuity control, not an exact reconstruction of original V-STaR semantics.
\item \textbf{\texttt{neutral\_rstr}}: a neutral wording control that preserves the same grounded event and output schema while removing physics-specific phrasing. It is best read as a templated ablation rather than as the paper's main semantic baseline.
\end{itemize}

The prompt families are not separate annotation pipelines. All three are derived from the same grounded event record, which fixes the event identity, temporal span, and spatial reference. For cross-family evaluation, \texttt{a\_when} and \texttt{a\_where} are therefore scored against the same targets in every family. The text field works slightly differently. The \texttt{physics} family uses the reference event description directly, while \texttt{neutral\_rstr} and \texttt{vstar\_like} use shorter family-appropriate semantic \texttt{a\_what} targets derived from that same record. This keeps the semantic target close to the answer style requested by the non-physics prompts while preserving the shared grounded event.

Figure~\ref{fig:prompt_families} shows an example of this shared-record design. The wording changes across prompt families, but the grounded event, temporal span, and spatial reference do not.

\subsection{Perturbation conditions}

Each base clip is evaluated under four input conditions:
\begin{itemize}
\item \textbf{Original}: the unmodified RGB video.
\item \textbf{Shuffled}: frames are randomly permuted while the set of frames is kept fixed.
\item \textbf{Ablated}: each frame is converted to greyscale and blurred to suppress color and fine texture while preserving coarse spatial structure.
\item \textbf{Frame-Masked}: half of the frames are replaced by black frames while video length and frame rate remain unchanged.
\end{itemize}

Each condition targets a different component of the visual signal. Shuffling removes temporal order while preserving the frame set. Ablation removes color and fine appearance detail while preserving coarse structure. Frame masking removes evidence intermittently while preserving the temporal structure. Together, these conditions let us ask whether a model depends on temporal order, appearance detail, or persistent visual evidence, instead of collapsing those effects into one original-input score.

\begin{figure*}[t]
\centering
\includegraphics[width=0.96\textwidth]{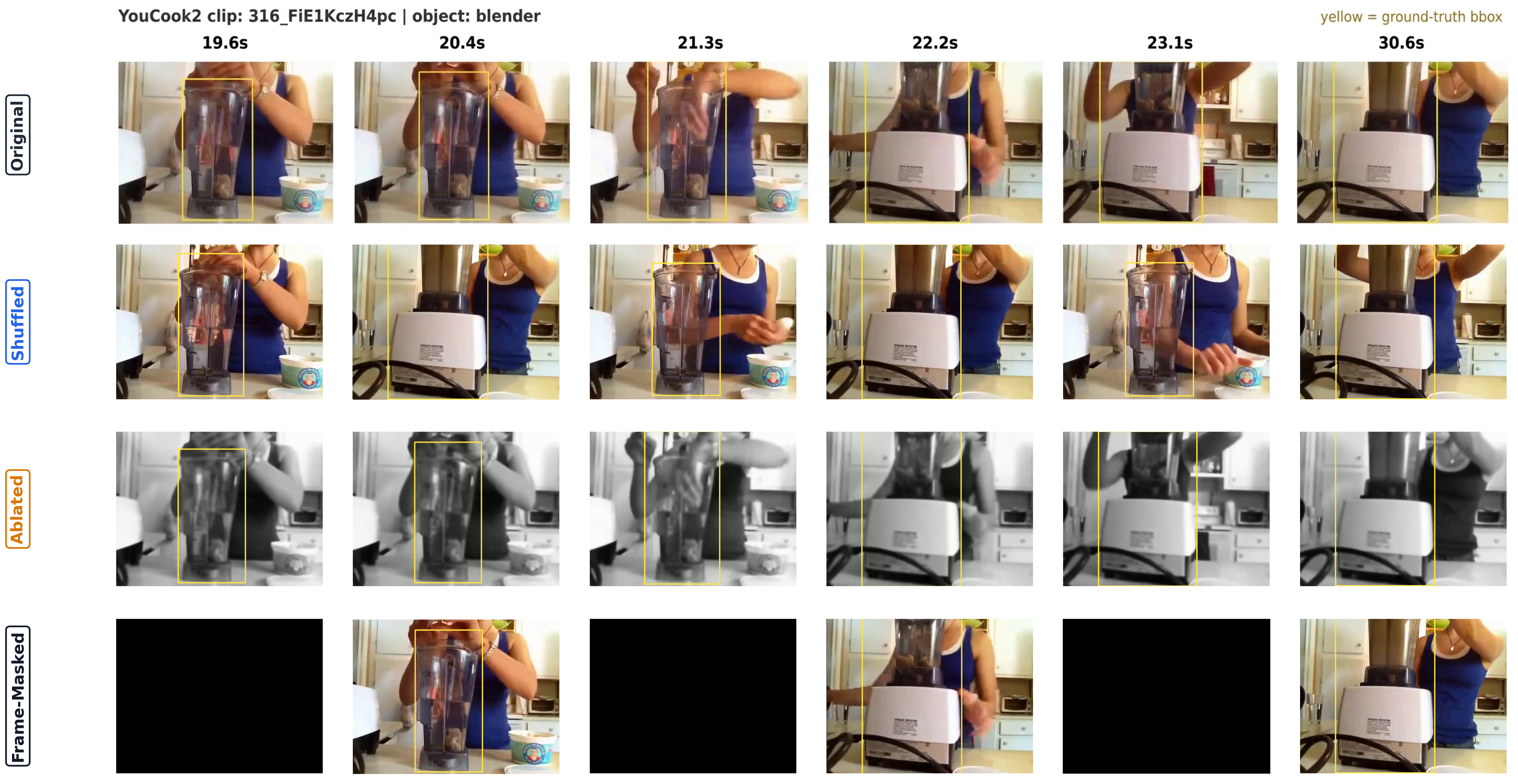}
\caption{The four input conditions applied to the same YouCook2 clip at matched timestamps. Yellow boxes show the ground-truth object reference across rows. Shuffling preserves the frame set while breaking temporal order; ablation suppresses color and fine texture while preserving coarse object extent; frame masking removes evidence intermittently.}
\label{fig:perturbation_storyboard}
\end{figure*}

Table~\ref{tab:perturbations} summarizes what each condition is meant to probe, and Figure~\ref{fig:perturbation_storyboard} shows the same clip under all four variants. The ablated row is especially useful because object extent remains visible while fine appearance cues are suppressed.

\begin{table}[t]
\centering
\small
\begin{tabular}{p{0.17\linewidth}p{0.30\linewidth}p{0.40\linewidth}}
\hline
Condition & Input change & Diagnostic role \\
\hline
Original & unmodified video & baseline performance \\
Shuffled & frame order permuted & temporal order sensitivity \\
Ablated & greyscale + blur & appearance dependence \\
Frame-Masked & 50\% frames replaced by black & robustness to missing evidence \\
\hline
\end{tabular}
\caption{The four evaluation conditions and the diagnostic role each is intended to probe. The grounded target is held fixed across conditions.}
\label{tab:perturbations}
\end{table}

\subsection{Metrics and reporting conventions}

We score the three output fields with grounded metrics. Let $\mathrm{Acc}$ be text accuracy for \texttt{a\_what}, $\mathrm{tIoU}$ be temporal intersection-over-union for \texttt{a\_when}, and $\mathrm{sIoU}$ be spatial intersection-over-union for \texttt{a\_where}. Following V-STaR~\cite{vstar}, we combine these three components with the Logarithmic Geometric Mean:
\begin{equation}
\begin{aligned}
\mathrm{LGM} =
-\frac{1}{3} \Big[
&\log(1-\mathrm{Acc}+\epsilon) \\
&+ \log(1-\mathrm{tIoU}+\epsilon) \\
&+ \log(1-\mathrm{sIoU}+\epsilon)
\Big]
\end{aligned}
\end{equation}
where $\epsilon$ is a small constant for numerical stability.

We use LGM for continuity with V-STaR, but interpret it together with the component scores rather than as a replacement for them. LGM is high only when all three components are strong. A model that achieves good semantic accuracy while failing at temporal or spatial localisation will still score poorly. In this setting, that behavior is intentional: strong \texttt{a\_what} performance should not hide weak \texttt{a\_when} or \texttt{a\_where} predictions.

There is one reporting detail worth stating explicitly. V-STaR applies an additional linear readability scaling to LGM~\cite{vstar}. Our pipeline reports LGM directly on the normalized component scale. This changes the displayed magnitude, but not the ranking or the perturbation ratios.

\subsection{Diagnostic indices}

We derive three indices from the four condition scores:
\begin{align}
\mathrm{SBI} &= 1 - (\mathrm{LGM}_{\mathrm{orig}} - \mathrm{LGM}_{\mathrm{shuf}}), \\
\mathrm{PRI} &= \frac{\mathrm{LGM}_{\mathrm{abl}}}{\mathrm{LGM}_{\mathrm{orig}}}, \\
\mathrm{SPI} &= \frac{\mathrm{LGM}_{\mathrm{mask}}}{\mathrm{LGM}_{\mathrm{orig}}}.
\end{align}

PRI and SPI are retention ratios relative to the original condition, so a value of $1$ means that the perturbed input matches the original-condition score. SBI is written on the same centered scale, with $1$ again meaning no change under temporal shuffling. Larger values therefore indicate greater robustness to the corresponding perturbation, while smaller values indicate loss under the perturbed input. We use these indices as descriptive diagnostics rather than standalone evidence: they are interpreted together with component deltas, baseline strata, validity transitions, and qualitative examples.

\subsection{Validation-aware scoring}

Model outputs are sometimes empty, malformed, or refusal-like. The evaluation pipeline counts those cases in the denominator and scores them as zero rather than silently dropping them. This matters for interpretation. If failed outputs were excluded, perturbation deltas would look artificially favorable for models that simply stop producing valid answers under harder settings.

%% file: sec/4_dataset.tex
\section{Dataset construction}

\subsection{Sources and scale}

The benchmark is constructed from four video sources that differ in viewpoint, event type, and annotation style: SSV2~\cite{ssv2}, YouCook2~\cite{youcook2}, HoloAssist~\cite{wang2023holoassist}, and Roundabout-TAU~\cite{lin2026taur1}. In total, the benchmark contains 1,560 base clips. Each base clip is expanded into four perturbation conditions, producing 6,240 scored video-condition pairs.

Table~\ref{tab:dataset_sources} shows the source breakdown. SSV2 and YouCook2 provide the largest portions of the data. HoloAssist contributes egocentric manipulation scenes in which hands and objects overlap under camera motion. Roundabout-TAU contributes overhead traffic footage where appearance cues are weak and motion structure matters more than texture.

\begin{table}[t]
\centering
\small
\begin{tabular}{lr}
\hline
Source & Base clips \\
\hline
SSV2 & 600 \\
YouCook2 & 600 \\
HoloAssist & 275 \\
Roundabout-TAU & 85 \\
\hline
Total & 1,560 \\
\hline
\end{tabular}
\caption{Source breakdown. Counts are for base clips before perturbation expansion.}
\label{tab:dataset_sources}
\end{table}

\subsection{Physics domains}

Every clip is assigned to one of six physics domains: Gravity, Fluids, Collisions, Deformation, Friction, and State Changes. The domain layer matters because it lets us ask not only whether a model fails, but where it fails. A model may perform well on State Changes because those events are temporally salient and semantically distinctive, yet still struggle on Friction or Collisions where localization is harder and the critical evidence can be brief or spatially small.

Table~\ref{tab:dataset_domains} shows the domain distribution across base clips. No domain is severely underrepresented, so per-domain analysis remains meaningful.

\begin{table}[t]
\centering
\small
\begin{tabular}{lr}
\hline
Domain & Base clips \\
\hline
Gravity & 248 \\
Fluids & 283 \\
Collisions & 296 \\
Deformation & 248 \\
Friction & 196 \\
State Changes & 289 \\
\hline
Total & 1,560 \\
\hline
\end{tabular}
\caption{Domain breakdown for the benchmark base clips.}
\label{tab:dataset_domains}
\end{table}

\subsection{Annotation pipeline}

Each base clip is converted into a common grounded record containing a reference event description, a temporal span, a bounding box, and a physics-domain label. The key design choice is that this record is prompt-family-agnostic. It defines which event should be grounded, when it happens, and where it occurs, before any particular query wording is chosen. Text annotations are produced with a local Qwen3.5-based generator~\cite{qwen35}, which rewrites source metadata and event windows into that shared event description. Temporal spans come from the source adapters and event windows rather than from a second prompt-generation stage.

Spatial annotations are generated on the original clip with GroundingDINO~\cite{groundingdino} and then reused across all four perturbation conditions. This keeps the target object reference fixed when the input is shuffled, ablated, or frame-masked. The prompt families are layered on top of the same grounded record. They do not regenerate temporal spans or spatial boxes. The semantic text target is handled more carefully: \texttt{physics} keeps the longer reference event description, while \texttt{neutral\_rstr} and \texttt{vstar\_like} derive shorter prompt-aligned semantic \texttt{a\_what} targets from the same grounded record.

We also performed a manual audit of approximately 120 generated records sampled across sources and physics domains. For each record, we checked whether the reference event was visible, whether the temporal span covered the relevant physical interaction, and whether the spatial target matched the relevant object or interaction region. The audit was not meant to make the benchmark fully human-verified; it was a sanity check for systematic pipeline errors and for cases where automatic supervision is fragile. The main ambiguous cases were egocentric clips, where hands and objects overlap, and traffic clips, where several vehicles may participate in the same interaction. We therefore interpret fine-grained spatial scores with this label noise in mind.

\subsection{Source adaptation and Roundabout selection}

The four sources are not simply merged. Each one is adapted into the same grounded annotation format and mapped into the six-domain taxonomy. SSV2 contributes short object-manipulation clips in which temporal order is often decisive. YouCook2 contributes longer procedural clips with extended state changes and fluid events. HoloAssist contributes first-person manipulation in which object visibility and grounding are harder because the camera moves with the actor.

Roundabout-TAU requires an additional filtering step. Much of the raw traffic footage shows ordinary circulation rather than a localized physical interaction. We therefore map the source event labels into our physics taxonomy and exclude normal traffic by default. Clips are retained only when the event can be grounded as a localized interaction, such as collision-like behavior or a physically meaningful maneuver conflict. This is also how we justify using Roundabout for the \textit{Collisions} domain: we do not treat every traffic clip as a collision example, only the subset whose event annotation corresponds to an interaction that can be temporally and spatially grounded.

\subsection{Prompt and perturbation reuse}

The query families are rendered automatically from the shared record rather than rewritten manually for each sample. This keeps supervision aligned across prompt families, but some prompts are terser or less natural than others. We therefore interpret cross-family results primarily through aggregate behavior over shared targets. Changes across prompt families or perturbation conditions reflect question formulation, visual evidence, and answer-style alignment rather than drift in the underlying grounded event.

%% file: sec/5_experiments.tex
\section{Experimental Setup}

\subsection{Models}

Table~\ref{tab:model_suite} lists the ten-model suite, spanning general-purpose VLMs, video-native models, a compact multimodal model, and recent open multimodal LLMs.

\begin{table*}[t]
\centering
\small
\begin{tabular}{lcc}
\hline
Model & Params & Family \\
\hline
VideoLLaMA3-7B & 7B & video-centric model \\
Qwen3-VL-8B-Instruct & 8B & general-purpose VLM \\
Molmo2-8B & 8B & multimodal LLM \\
Qwen2.5-VL-7B-Instruct & 7B & general-purpose VLM \\
Gemma4-26B-A4B-IT & 26B & multimodal LLM \\
MiniCPM-o~2.6 & 2.6B & compact multimodal model \\
Qwen3-VL-8B-Thinking & 8B & thinking-variant VLM \\
InternVideo2.5-Chat-8B & 8B & video-native model \\
InternVL3.5-8B & 8B & image-centric VLM \\
Qwen3.5-9B & 9B & native multimodal LLM \\
\hline
\end{tabular}
\caption{Ten-model suite spanning general-purpose VLMs, video-native models, compact multimodal models, and open multimodal LLMs.}
\label{tab:model_suite}
\end{table*}

We evaluate Qwen2.5-VL~\cite{qwen25vl}, Qwen3-VL and Qwen3-VL-Thinking~\cite{bai2025qwen3vl}, VideoLLaMA3~\cite{videollama3}, InternVideo2.5~\cite{internvideo25}, InternVL3.5~\cite{wang2025internvl35}, MiniCPM-o~\cite{minicpm}, Qwen3.5~\cite{qwen35}, Gemma~4~\cite{gemma-4-26b-a4b-it}, and Molmo2~\cite{clark2026molmo2}. The suite gives us useful contrasts: two Qwen generations, two video-native models, one image-centric VLM, a compact multimodal model, and several recent open multimodal LLMs.

\subsection{Inference setup}

All models return one JSON object containing \texttt{a\_what}, \texttt{a\_when}, and \texttt{a\_where}. We evaluate all four perturbation conditions, and cross-family comparisons use the full completed model set. Frame budgets follow native presets rather than a single shared budget because the benchmark is meant to measure model behavior in its normal operating mode. The most informative comparisons are therefore within-model changes across prompt families and perturbations, not perfectly matched frame counts across architectures.

\subsection{Evaluation and analysis protocol}

For each sample, the prediction is compared to the reference annotation using text accuracy, temporal IoU, spatial IoU, and LGM. Condition-level means are computed first, and SBI, PRI, and SPI are derived from those means. Missing, malformed, or refusal-like outputs are scored as zero and kept in the denominator. Otherwise the benchmark would overstate models that simply stop returning usable answers under harder conditions.

The main original-condition physics comparison uses all ten integrated models. Prompt-family comparisons use the same suite, with shared \texttt{a\_when} and \texttt{a\_where} targets and prompt-aligned non-physics \texttt{a\_what} targets derived from the same event record.

We interpret perturbation behavior with analyses beyond the aggregate indices. For selected models and prompt families, we use per-sample $\Delta$LGM, bootstrap confidence intervals, sign tests, baseline-stratified summaries, leave-one-dataset-out checks, and qualitative examples to interpret gains and losses.

%% file: sec/6_results.tex
\section{Results}

\subsection{Main physics benchmark}

Table~\ref{tab:main_results} reports original-condition grounded metrics in the \texttt{physics} prompt family together with the three perturbation indices.

\begin{table*}[t]
\centering
\small
\begin{tabular}{lccccccc}
\hline
Model & Acc & tIoU & sIoU & LGM & SBI & PRI & SPI \\
\hline
VideoLLaMA3 & 0.319 & 0.547 & 0.042 & \textbf{2.634} & 0.959 & 0.926 & 1.032 \\
Qwen3-VL & \textbf{0.687} & 0.183 & 0.056 & 2.551 & 1.076 & 1.231 & \textbf{1.337} \\
Molmo2 & 0.548 & 0.322 & 0.023 & 2.515 & 1.326 & 1.087 & 0.974 \\
Qwen2.5-VL & 0.378 & \textbf{0.560} & \textbf{0.073} & 2.399 & 1.076 & 1.073 & 1.091 \\
Gemma4 & 0.481 & 0.455 & 0.034 & 2.150 & 1.173 & 1.047 & 0.862 \\
MiniCPM-o~2.6 & 0.232 & 0.373 & 0.033 & 1.396 & 0.947 & 0.946 & 0.706 \\
Qwen3-VL-Thinking & 0.635 & 0.126 & 0.051 & 1.332 & 1.106 & 1.304 & 1.132 \\
InternVideo2.5 & 0.189 & 0.285 & 0.027 & 1.223 & 1.290 & 1.064 & 0.705 \\
InternVL3.5 & 0.393 & 0.263 & 0.036 & 0.828 & 1.302 & 1.058 & 1.001 \\
Qwen3.5 & 0.310 & 0.323 & 0.037 & 0.766 & 1.309 & 1.093 & 0.965 \\
\hline
\end{tabular}
\caption{Physics-prompt results averaged over the four sources. Acc, tIoU, sIoU, and LGM are original-condition scores; SBI, PRI, and SPI summarize perturbation response. Values above $1$ mean the perturbed condition outscored the original. Models are ordered by LGM values.}
\label{tab:main_results}
\end{table*}

Table~\ref{tab:main_results} shows the same failure pattern that motivated V-STaR: strong semantic accuracy does not guarantee equally strong temporal or spatial grounding. Qwen3-VL leads Acc, Qwen2.5-VL leads tIoU and sIoU, and VideoLLaMA3 leads LGM. These models arrive there in different ways: VideoLLaMA3 is strongest temporally, Qwen3-VL is strongest semantically, and Qwen2.5-VL is the most balanced on temporal and spatial localization. Molmo2 joins this top group with a distinct perturbation profile. Spatial grounding remains the clearest weakness, with no model exceeding 0.073 mean sIoU.

\subsection{Prompt-family changes and V-STaR-style prompting}

Table~\ref{tab:cross_family_lgm} reports original-condition LGM across prompt families on the full ten-model set.

\begin{table*}[t]
\centering
\small
\begin{tabular}{lccc}
\hline
Model & \texttt{physics} & \texttt{vstar\_like} & \texttt{neutral\_rstr} \\
\hline
VideoLLaMA3 & 2.634 & 2.171 & 1.181 \\
Qwen3-VL & 2.551 & 0.116 & 0.114 \\
Molmo2 & 2.515 & 1.384 & 0.866 \\
Qwen2.5-VL & 2.399 & 1.058 & 0.750 \\
Gemma4 & 2.150 & 0.945 & 1.315 \\
MiniCPM-o~2.6 & 1.396 & 1.824 & 0.515 \\
Qwen3-VL-Thinking & 1.332 & 0.242 & 0.102 \\
InternVideo2.5 & 1.223 & 0.839 & 0.304 \\
InternVL3.5 & 0.828 & 0.448 & 0.297 \\
Qwen3.5 & 0.766 & 0.524 & 0.214 \\
\hline
Mean over 10-model set & 1.779 & 1.139 & 0.547 \\
\hline
\end{tabular}
\caption{Original-condition LGM across prompt families on the full ten-model set. Non-physics rows use prompt-aligned semantic \texttt{a\_what}; \texttt{a\_when} and \texttt{a\_where} remain shared across families. Models are ordered by physics LGM values.}
\label{tab:cross_family_lgm}
\end{table*}

Table~\ref{tab:cross_family_lgm} and Figure~\ref{fig:prompt_families} show the same ordering: \texttt{physics} is strongest overall, \texttt{vstar\_like} sits between it and \texttt{neutral\_rstr}, and mean LGM falls from 1.779 to 1.139 to 0.547, making \texttt{vstar\_like} the main non-physics comparison.

Cross-family shifts are model-specific rather than uniform. VideoLLaMA3 remains strong under \texttt{vstar\_like}, and MiniCPM-o~2.6 is the clearest positive case: its \texttt{vstar\_like} score exceeds its \texttt{physics} score. Molmo2 also recovers meaningfully under \texttt{vstar\_like}, while Gemma4 is the cleanest \texttt{neutral\_rstr} case. The strongest negative cases are Qwen3-VL and Qwen3-VL-Thinking, which both drop sharply outside \texttt{physics}. Qwen3.5 follows the same pattern at a lower level. High performance under physics-framed queries therefore does not necessarily transfer to alternative semantic~formulations.

\subsection{Interpreting perturbation indices}

By construction, PRI and SPI compare ablated and masked performance to the original condition, and SBI compares shuffled performance to the original on the same centered scale. Values above $1$ mean that the perturbed input outscored the original, but they should be read as diagnostics rather than blanket robustness. Across the analyses, low- and mid-baseline rows are more likely to improve than high-baseline rows.

\begin{table}[t]
\centering
\small
\begin{tabular}{@{}p{0.33\linewidth}@{\hspace{0.4em}}p{0.13\linewidth}@{\hspace{0.4em}}p{0.14\linewidth}@{\hspace{0.4em}}p{0.18\linewidth}@{}}
\hline
Model / family & Perturb. & Mean $\Delta$LGM & Reading \\
\hline
Qwen3-VL / physics & ablated & +0.589 & strong-model gain \\
Molmo2 / physics & shuffled & +0.326 & mid-baseline temporal gain \\
Gemma4 / neutral\_rstr & shuffled & +0.347 & control-family gain \\
InternVL3.5 / physics & shuffled & +0.302 & low-baseline gain \\
\hline
\end{tabular}
\caption{Representative positive-response cases. Positive $\Delta$LGM is descriptive, not a general robustness claim.}
\label{tab:perturbation_cases}
\end{table}

Figure~\ref{fig:perturbation_storyboard} and Table~\ref{tab:perturbation_cases} show why the sign alone is not enough under \texttt{physics}. The same sign can arise in different regimes, so positive $\Delta$LGM should not be read as one uniform notion of robustness. Qwen3-VL and VideoLLaMA3 make the contrast clear: their original-condition LGM values are close, but Qwen3-VL often improves under perturbation whereas VideoLLaMA3 is flatter or negative on shuffled and ablated inputs.

The case-level analysis points to a more specific failure pattern. Perturbations often rescue rows that were already weak on the original input, for example by making a temporal guess easier or by reducing distracting appearance detail. They usually do not improve rows where the model was already well grounded. In those stronger rows, shuffling tends to remove useful order information and masking removes evidence the model was using. This is why we avoid describing above-$1$ indices as better physical reasoning. They are better read as evidence sensitivity: the model's score changes when order, appearance, or frame availability is changed, but the mechanism depends on the original baseline and on which component, \texttt{a\_what}, \texttt{a\_when}, or \texttt{a\_where}, moved.

\subsection{Domain-level findings}

Domain leaders vary by domain: Qwen3-VL leads Gravity and Friction, VideoLLaMA3 leads Fluids, and Qwen2.5-VL leads Collisions, Deformation, and State Changes. There is no winner-take-all leader. Domain-level LGM changes across models, the underlying limitation does not: spatial grounding remains weak even when a model leads a domain.

%% file: sec/7_limitations.tex
\section{Limitations}

\textbf{Automatic supervision.} Event descriptions, temporal spans, and spatial boxes are produced automatically rather than fully verified by human annotators. That gives the benchmark scale, but it also introduces noise. Our manual audit of approximately 120 records helped catch obvious pipeline errors, but it is not a substitute for exhaustive human verification. The hardest cases are egocentric and traffic videos, where the grounded target can be ambiguous even for a human reader. A larger human validation pass would improve confidence in the labels, especially for spatial grounding.

\textbf{Annotation-style bias.} The text annotations are generated with an LLM-based stage, and one evaluated model family is closely related to that generator. We therefore use shorter non-physics semantic \texttt{a\_what} targets derived from the shared event record instead of reusing the longer physics-style answers verbatim. That keeps cross-family semantics closer to the answer style requested by the prompts, but the non-physics targets remain deterministic derivations rather than a manually curated multi-family gold set. Large cross-family drops should therefore not be over-attributed to any single cause: instruction-following preferences, answer-style priors, and deeper failures in grounded video reasoning can all contribute.

\textbf{Source imbalance.} The benchmark is reasonably balanced by domain, but not by source. Roundabout-TAU contributes 85 clips, compared with 600 each from SSV2 and YouCook2, so source-level findings should be read cautiously.

\textbf{Scope of the task.} The benchmark evaluates grounded event understanding: what happened, when it happened, and where it happened. It does not directly test causal forecasting, counterfactual physics, or long-horizon planning. A model that localises a collision correctly can still fail to predict what follows or to reason about alternative physical outcomes.

\textbf{Diagnostic, not causal, interpretation.} Perturbations expose behavior that original-condition scores hide, but they do not identify a single underlying cause. A gain under ablation can reflect reduced distractors, weak original baselines, or a cleaner temporal guess, so we treat perturbations as diagnostics, not causal explanations.

%% file: sec/8_conclusion.tex
\section{Conclusion}

We introduced a grounded benchmark for physical video understanding that extends the \textit{what--when--where} diagnostic idea of V-STaR~\cite{vstar} to physics-focused video, multiple prompt families, and controlled perturbations built on a shared grounded event record. The experiments support three conclusions: \texttt{physics} is strongest overall, \texttt{vstar\_like} is the strongest non-physics comparison, \texttt{neutral\_rstr} is better read as a harder control, and prompt-family robustness is selective rather than universal. Spatial grounding remains the weakest component, so answer accuracy alone still hides too much behavior. A model can still recover the event label while failing to ground it in time or space, and prompt-family changes and input perturbations make that failure easier to see.